\documentclass[lettersize,journal]{IEEEtran}
\usepackage{amsmath,amsfonts}
\usepackage{algorithmic}
\usepackage{algorithm}
\usepackage{array}
\usepackage[caption=false,font=normalsize,labelfont=sf,textfont=sf]{subfig}
\usepackage{textcomp}
\usepackage{stfloats}
\usepackage{url}
\usepackage{verbatim}
\usepackage{graphicx}
\usepackage{cite}
\usepackage{orcidlink}
\usepackage{diagbox}
\usepackage{multirow}
\usepackage{cleveref}

\begin{document}
\title{Deep Learning for Precision Agriculture: Post-Spraying Evaluation and Deposition Estimation}
 \author{Harry Rogers \orcidlink{0000-0003-3227-5677} , Tahmina Zebin \orcidlink{0000-0003-0437-0570}, Grzegorz Cielniak \orcidlink{0000-0002-6299-8465}, Beatriz De La Iglesia \orcidlink{0000-0003-2675-5826}, Ben Magri 
 \thanks{This work is supported by the Engineering and Physical Sciences Research Council [EP/S023917/1]. This work is also supported by Syngenta as the Industrial partner.}
 \thanks{Harry Rogers and Beatriz De La Iglesia, University of East Anglia, United Kingdom, Harry.Rogers@uea.ac.uk, B.Iglesia@uea.ac.uk, 
 Tahmina Zebin, Brunel University London, United Kingdom, tahmina.zebin@brunel.ac.uk,
 Grzegorz Cielniak, University of Lincoln, United Kingdom, gcielniak@lincoln.ac.uk, Ben Magri, Syngenta, United Kingdom, ben.magri@syngenta.com.}}

\markboth{Journal of \LaTeX\ Class Files,~Vol.~14, No.~8, August~2021}%
{Shell \MakeLowercase{\textit{et al.}}: A Sample Article Using IEEEtran.cls for IEEE Journals}


\maketitle

\begin{abstract}
Precision spraying evaluation requires automation primarily in post-spraying imagery. In this paper we propose an eXplainable Artificial Intelligence (XAI) computer vision pipeline to evaluate a precision spraying system post-spraying without the need for traditional agricultural methods. The developed system can semantically segment potential targets such as lettuce, chickweed, and meadowgrass and correctly identify if targets have been sprayed. Furthermore, this pipeline evaluates using a domain-specific Weakly Supervised Deposition Estimation task, allowing for class-specific quantification of spray deposit weights in $\mu$L. Estimation of coverage rates of spray deposition in a class-wise manner allows for further understanding of effectiveness of precision spraying systems. Our study evaluates different Class Activation Mapping techniques, namely AblationCAM and ScoreCAM, to determine which is more effective and interpretable for these tasks. In the pipeline, inference-only feature fusion is used to allow for further interpretability and to enable  the automation of precision spraying evaluation post-spray. Our findings indicate that a Fully Convolutional Network with an EfficientNet-B0 backbone and inference-only feature fusion achieves an average absolute difference in deposition values of 156.8 $\mu$L across three classes in our test set. The dataset curated in this paper is publicly available at https://github.com/Harry-Rogers/PSIE

\end{abstract}


\section{Introduction}
\IEEEPARstart{A}{utomated} precision spraying in precision agriculture requires efficient evaluation post-spraying to locate and quantify spray deposits. Current methods involve human intervention, leading to increased costs and potential inaccuracies. The predominant techniques utilized are Water Sensitive Papers (WSPs) and tracers, each accompanied by its own set of limitations. While these methods have been instrumental in assessing spray deposition, they fall short in providing automated results. Moreover, the reliance on human intervention introduces the possibility of subjective errors and delays in the decision-making processes. Hence, there is a pressing need for innovative solutions that can autonomously evaluate spray deposition accurately in a timely manner.

We have previously worked on spraying evaluation post-spray without the use of traditional methods with great
success within the classification of pre- or post-spraying \cite{CASE, TAROS, NEURAL_JOURNAL}. However, there were limitations such as knowing what object is sprayed and how much has been deposited. To address these  issues, using the same dataset, semantic segmentation annotations are added to generate a ground truth for each object class. There are 7 classes: background, lettuce, chickweed, meadowgrass, sprayed lettuce, sprayed chickweed, and sprayed meadowgrass. The dataset has been made publicly available  alongside this paper. Further information on this dataset is in \Cref{sec3:data}. Additionally, a domain-specific Weakly Supervised Deposition Estimation (WSDE) task has been added to the dataset. To estimate deposition values, Class Activation Maps (CAMs), that highlight regions of interest from Deep Learning models using the last convolutional layer, are combined with the Deep Learning model prediction itself to create deposition values in a class-wise manner.  This is not possible with traditional methods.

For this new dataset and task, an eXplainable Artificial Intelligence (XAI) pipeline is proposed to complete semantic segmentation of crops and weeds pre- and post-spraying. The developed pipeline can be used to evaluate precision sprayers without the need for traditional manual agricultural methods. To improve model accuracy, inference-only feature fusion has been developed combining auxiliary outputs with the traditional output. 

In the XAI pipeline two CAM methodologies are compared and evaluated using CAM metrics to understand how representative a CAM is of model predictions. Inference-only feature fusion is compared to a baseline to identify if it could be more interpretable. We are assessing if  more of the early stage layers within Deep Learning models, included in the inference-only feature fusion, could help understand model predictions.

The core contributions of this paper are:
\begin{itemize}
    \item An automated process to create deposition values in a class wise manner without WSPs or tracers.
    \item Inference-Only feature fusion to improve from baseline model in interpretability and segmentation metrics.
    \item An open dataset for post-spraying evaluation.
\end{itemize}

The remainder of this paper is organized as follows. \Cref{sec:lit} introduces related work on precision spraying systems and how systems are evaluated post-spray, and XAI methods for generating and evaluating CAMs. \Cref{sec3:data} introduces the AI semi-automatic annotation methodology and relevant coverage rate information for the dataset. \Cref{sec4:pipe} provides details of the XAI pipeline workflow, Deep Learning architectures utilized, and evaluation metrics for segmentation, CAMs, and the WSDE task. The results of the segmentation models, CAMs, and WSDE scores are reported in \Cref{results}. Finally, conclusions and future work are presented in \Cref{sec:future}.

\section{Related Work}
\label{sec:lit}
We review current precision spraying systems and evaluation methods to demonstrate the need for an automated method.  CAMs for semantic segmentation and evaluation are also investigated.

\subsection{Precision Spraying evaluation}
Precision spraying in agriculture is crucial for the sustainable and efficient application of chemicals over large areas of land. Sprayers must be precisely calibrated to ensure that chemicals are accurately deposited on target areas, thereby minimizing waste and environmental impact. According to the 2019 European Union (EU) Green Deal \cite{european_commission_2019}, modern precision sprayers will need to undergo further regulatory assessment to confirm their ability to minimize chemical usage by ensuring accurate application. Therefore, developing robust methodologies to evaluate the effectiveness of these systems, particularly in terms of accurately landing spray deposits on desired targets, is of paramount importance. Currently there are two primary approaches for precision spraying evaluation. Namely, traditional agricultural or sprayer specific. Proposed precision sprayers typically use one of these to evaluate post-spraying.

One popular approach within traditional agricultural methods, the most common within the literature, is WSPs. WSPs are yellow pieces of paper that are used as targets for precision sprayers. When sprayed WSPs change color and make deposition values possible with computer vision \cite{inbook}. Many types of aerial and ground precision spraying systems have been developed and evaluated with WSPs. Applications within the literature vary widely and include  weed spraying in corn fields, cabbage fields, and cereal fields \cite{10075447, agronomy12102551, GONZALEZDESOTO2016165, LI2023107755}, pest control \cite{10.3389/fpls.2022.1042769}, disease detection in potatoes \cite{rs12244091, FAROOQUE2023100073}, orchard tree spraying \cite{9197556, cai2019design, Seol2022}, and vineyard spraying \cite{PARTEL2021106556, electronics10172061}. Testing can also be for systems that are static and are under development \cite{make6010014, RAJA202331}. Aerial spraying has been primarily explored with usage of WSPs \cite{drones6120383, wang2021smart, https://doi.org/10.1002/ps.5321, app9020218, Martinez-Guanter2020}. Despite this, WSPs have several drawbacks. Firstly, human intervention is required to place and retrieve the WSPs for analysis post-spraying, making the process labor intensive. Secondly, the texture of WSPs differs from the actual targets, potentially leading to differences between actual and estimated spray deposits. Lastly, no deployed spraying system can perfectly replicate the same spray deposit therefore deposits on WSPs will not be the same as the deposits on real targets.

Tracers are another traditional agricultural method developed to improve upon WSPs. They are typically dyes that color the chemical used for spraying. Tracers make the location of deposits clearly visible when sprayed on target or non-targets in error. This means that spray deposits from each system can be evaluated directly. Similarly to WSPs, tracers have been explored within a wide variety of applications. For example, pesticide spraying in Maize fields, orchards, and rice fields \cite{agriculture13030691, agronomy12102509, gao2019water}. Weed control has also been explored in field and in controlled greenhouse environments \cite{RAJA2020257, RAJA202331, liu2021development, OZLUOYMAK2022107134, Xun2023}. However, not all systems can use the same tracer, as different tracers may be incompatible with specific chemical applications. For example, Gao et al. \cite{gao2019water} use Allura red food dye as a tracer for pesticide spraying, but the drawback of the developed system is that to analyse the spray deposits, the target must be harvested and then tested. Zheng et al. \cite{agriculture13030691} use Rhodamine dye for pesticide spraying, whereas Liu et al. \cite{agronomy12102509} use Tartrazine dye for pesticide spraying. Thus, the type of tracer is dependent on what exact chemical is being sprayed as there are no generalisable tracers that work with all types of chemical and spraying application. 

There are some systems in the literature that are evaluated without traditional agricultural methods. These are sprayer specific methodologies. Some precision spraying systems are evaluated with human intervention, where humans manually count the number of weeds sprayed \cite{PARTEL2019339, s20247262}. Some other systems calculate the volume of chemical sprayed \cite{s22249723}. Some systems just assume that they hit the target, \cite{10.1371/journal.pone.0283801,salazargomez2021practical}. Furthermore, some state the evaluation of spraying post-spray is future work \cite{Salas2024}. However, these evaluation methods create ambiguity as it is unclear how effective precision sprayers truly are. Furthermore, these type of evaluation systems are typically system specific, and do not generalise well to other systems. 

From this review, it can be stated that precision spraying systems need an automated process that can evaluate post-spraying without the usage of WSPs or tracers. The automated method also needs to generate deposition values in a class-wise manner to improve from traditional methods. Therefore, a XAI pipeline has been developed that can do so by using Deep Learning and inference-only feature fusion.  

\subsection{XAI methodologies}
XAI has several methodologies focused on identifying regions of interest in images generated by Deep Learning model predictions. These approaches, utilizing either gradient or activations from the predictions, concentrate on the model's final layer before classification to compute CAMs. These representations are typically visualized for interpretation. Gradient-based methods like GradCAM, GradCAM++, and FullGrad have shown effectiveness in visualizing these regions of interest \cite{Selvaraju_2019,chattopadhay2018grad,fullgrad}. However, recent trends are moving towards gradient-free techniques to better accommodate models with gradients that are either negative or non-differentiable, typically used within Deep Learning for more complex tasks than classification such as segmentation.

A leading gradient-free approach, AblationCAM, evaluates the significance of activations by measuring the impact of their removal on the model output \cite{ramaswamy2020ablation}. This method has shown excellent performance over GradCAM for Convolutional Neural Networks (CNNs) trained on the ImageNet dataset \cite{5206848}. Other gradient-free CAM strategies like EigenCAM, which calculates the principal component analysis of network activations, has set new performance benchmarks on ImageNet whilst being more efficient \cite{bany2021eigen}. ScoreCAM, another approach, employs a unique two-step process that combines activation maps as masks on the original image to generate a combined output \cite{wang2020score}. These are the leading three CAM methods using activations only.

To evaluate CAM effectiveness, several metrics are used, including Deletion, Insertion, and a weakly supervised task \cite{rise}. Deletion quantifies the change in confidence when different regions of an image are removed by setting pixels to 0. Insertion measures the confidence change where the region of interest is added to an image with no surrounding context. The literature shows differing types of methodologies of Deletion and Insertion with altering perturbation methods. Specifically, the Most Relevant First (MoRF) imputation strategy introduced by Tomsett et al. \cite{tomsett2019sanity} involves removing the most relevant pixels first as these regions will be used in a WSDE task. The classical versions of Deletion and Insertion mentioned previously where 0 values are used, is used as opposed to blurring by Rong et al. \cite{rong2022consistent}. This is to get a true baseline for the comparison of CAM methods. In our weakly supervised task presented in this paper,  CAMs are converted to key points, taking inspiration from Ryou and Perona's work \cite{ryou2021weakly}, to then be evaluated using the pointing game from Zhang et al. \cite{Zhang2018}. These are further explained in \Cref{sec4:pipe}.

From the literature, this paper will compare AblationCAM and ScoreCAM using Deletion and Insertion with our collected precision spraying evaluation dataset. EigenCAM could be used but as it is class indiscriminate this will visualize poorly with multiple classes in each image against ScoreCAM and AblationCAM, which have class specific information. After evaluating the effectiveness of both CAM methods and identification of a best method, a domain-specific WSDE as described in \Cref{sec4:pipe} will be completed. Inference-only feature fusion is proposed to identify if adding earlier information from Deep Learning models with CAM generation is more interpretable than just the traditional output. Further details on the Semantic Segmentation Architectures and inference-only feature fusion is in \Cref{sec4:pipe}.

\section{Dataset}
\label{sec3:data}
The dataset is publicly available at \url{https://github.com/Harry-Rogers/PSIE}. 

\subsection{XY spot sprayer}
The dataset contains images of trays of 4 evenly spaced lettuces with randomly sown chickweed and meadowgrass that are commonly found in fields.

The precision sprayer evaluated in this paper is called the XY spot sprayer. Syngenta developed this as an experimental Agri-robot precision spraying system. The system uses an XY gantry system with an adjustable floor to spray at differing heights and locations. Added to the system is a Canon 500D camera to capture images pre- and post-spraying. Syngenta recommended a spraying height of 30 cm from the tray bed with a 3-bar pressure and a spray time of 8 milliseconds.

The dataset is made up of 176 images of trays that contain lettuce, chickweed, and meadowgrass. A tray was placed into the precision spraying system and an image was taken from the attached camera. After image capture the precision sprayer, controlled by an expert, sprayed the target chickweed and meadowgrass once. After spraying was complete another image was captured. Therefore, this dataset looks at the accuracy of the targeting within the precision spraying system and can be used to identify coverage and deposition values. All trays within the dataset were stored in a greenhouse with misting on, this means that lettuces, chickweed, or meadowgrass may look sprayed but are just wet and are not actually sprayed by the precision sprayer. This adds a layer of complexity to the data which may resemble real world situations and may ensure precision spraying deposits can be identified in real world scenarios.

\subsection{Semi-automatic segmentation annotation}
\begin{figure}[h]
  \centering
  \subfloat[Sprayed Chickweed.]{\includegraphics[height=0.4\linewidth]{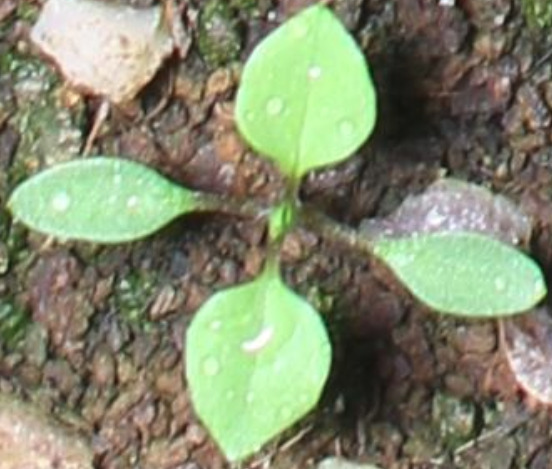}\label{example_drop}}
  \hspace{0.2cm}
  \subfloat[Highlighted droplets.]{\includegraphics[height=0.4\linewidth]{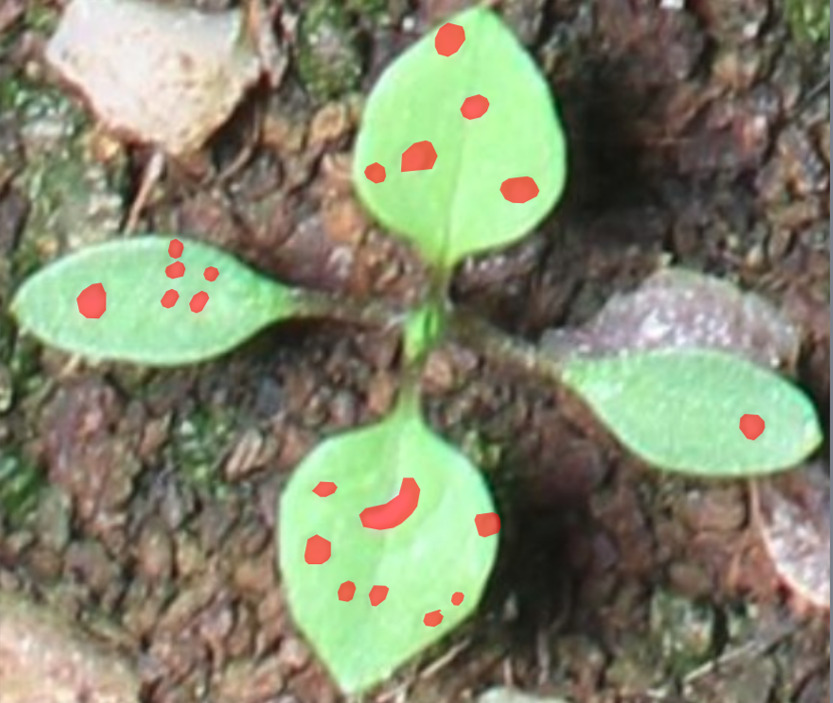}\label{red_drop}}
  \caption{Example of a single spray deposit.}
  \label{Single_deopsit}
\end{figure}

The collected data was annotated for a semantic segmentation task as an efficient way to label the data collected. Due to the granularity of the images as shown in \Cref{Single_deopsit}, segmentation of individual droplets is difficult or not possible for smaller droplets, but segmentation of sprayed objects for this task is more effective.

The classes include background, lettuce, chickweed, meadowgrass, sprayed lettuce, sprayed chickweed, and sprayed meadowgrass. A background class has been added as it allows for the semantic segmentation model to assign a class to all pixels and generalize better. From the ground truth labels several statistics can be identified. For example, there are a total of 9542 annotations with 398 instances of Lettuces, 3098 instances of Chickweed, 3529 instances of Meadowgrass, 310 instances of sprayed Lettuces, 1732 instances of Sprayed Chickweed, 299 instances of Sprayed Meadowgrass, and 176 background instances. The data is split into a 160 images for training, and 16 images for test. As images are pre- and post-spraying there can be images where all class instances exist.

\begin{table}[h]
\caption{Coverage Rate}
\centering
\label{tab:coverage}
\begin{tabular}{lrrrr}
Class & \begin{tabular}[c]{@{}r@{}}Coverage \\ {[}$CM^{2}${]}\end{tabular} & \begin{tabular}[c]{@{}r@{}}Average \\ Coverage\\  {[}$CM^{2}${]}\end{tabular} & Miss (\%) & Hit (\%) \\ \hline
Lettuce & 9448.2 & 107.3 & 6.9 & \textbf{93.1} \\ 
Chickweed & 4923.9 & 55.9 & 23.9 & \textbf{76.1} \\ 
Meadowgrass & 1856.5 & 21.1 & \textbf{83.1} & 16.9 \\ 
\end{tabular}
\end{table}

From the dataset spray deposition and targeting statistics can be identified. Shown in \Cref{tab:coverage} are the coverage values for the entire dataset with the hit rate and miss rate as a percentage. It can be seen that on average lettuce covers an area of 41.1 \textit{$CM{^2}$}, chickweed covers 19.5 \textit{$CM{^2}$}, and meadowgrass covers 13.3 \textit{$CM{^2}$}. With the dataset collected lettuce is hit 93.1\% of the time (in error as it should not be sprayed), chickweed is hit 76.1\% of the time, and meadowgrass is hit 16.9\% of the time. These are calculated using the number of instances of sprayed objects in the given class.

The dataset has an unbalanced class representation, like real-world scenarios where systems are designed to spray weeds in general. It also includes misfires and hits on non-target lettuces, mimicking the challenges a real system would face, such as inertia and other real-world issues.

\begin{figure}[htbp]
    \centering
    \includegraphics[height=275px]{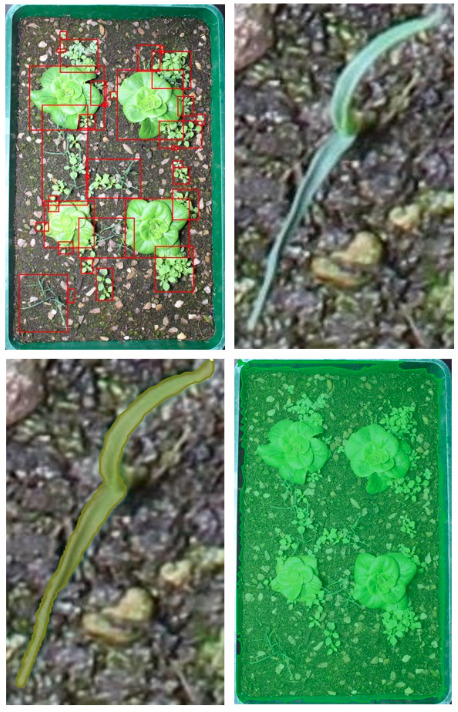}
    \caption{Semi automatic process for segmentation annotation using SAM.}
    \label{fig:auto}
\end{figure}

Data was annotated semi-automatically as semantic segmentation labels are labor intensive. First a human labeler created bounding boxes around the lettuces and clusters of weeds regardless of type as shown in \Cref{fig:auto} in the top left. Clusters are defined as any weeds that overlap with each other. 

\begin{figure}[h]
  \centering
  \subfloat[RGB Image.]{\includegraphics[width=0.3\linewidth]{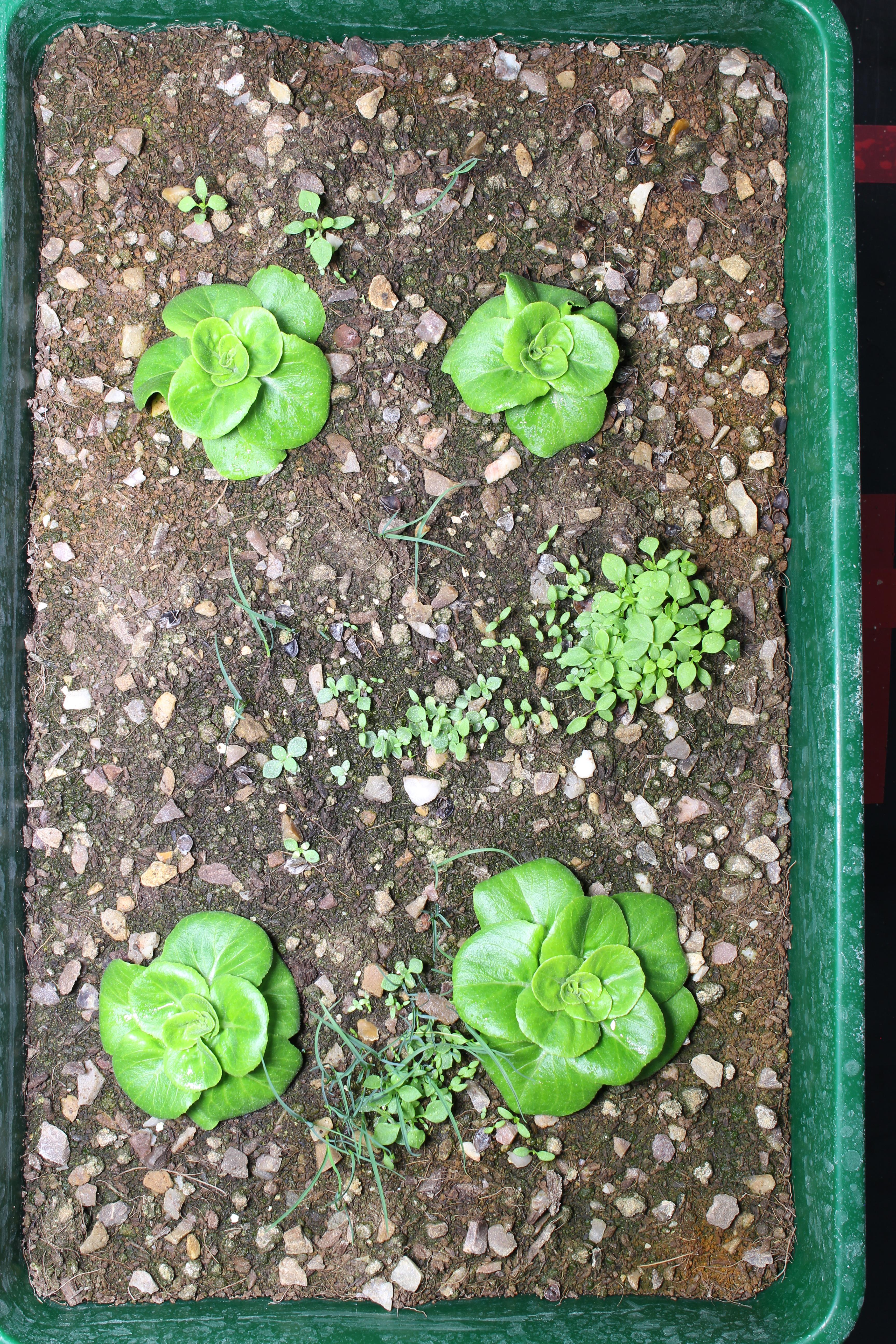}\label{fig2a}}
  \hspace{0.1cm}
  \subfloat[Ground Truth.]{\includegraphics[width=0.3\linewidth]{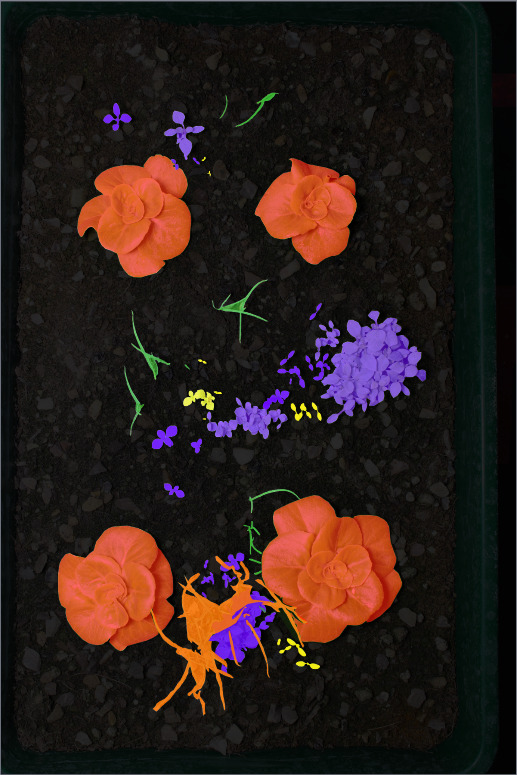}\label{fig2b}}
    \hspace{0.1cm}
  {\includegraphics[width=0.3\linewidth]{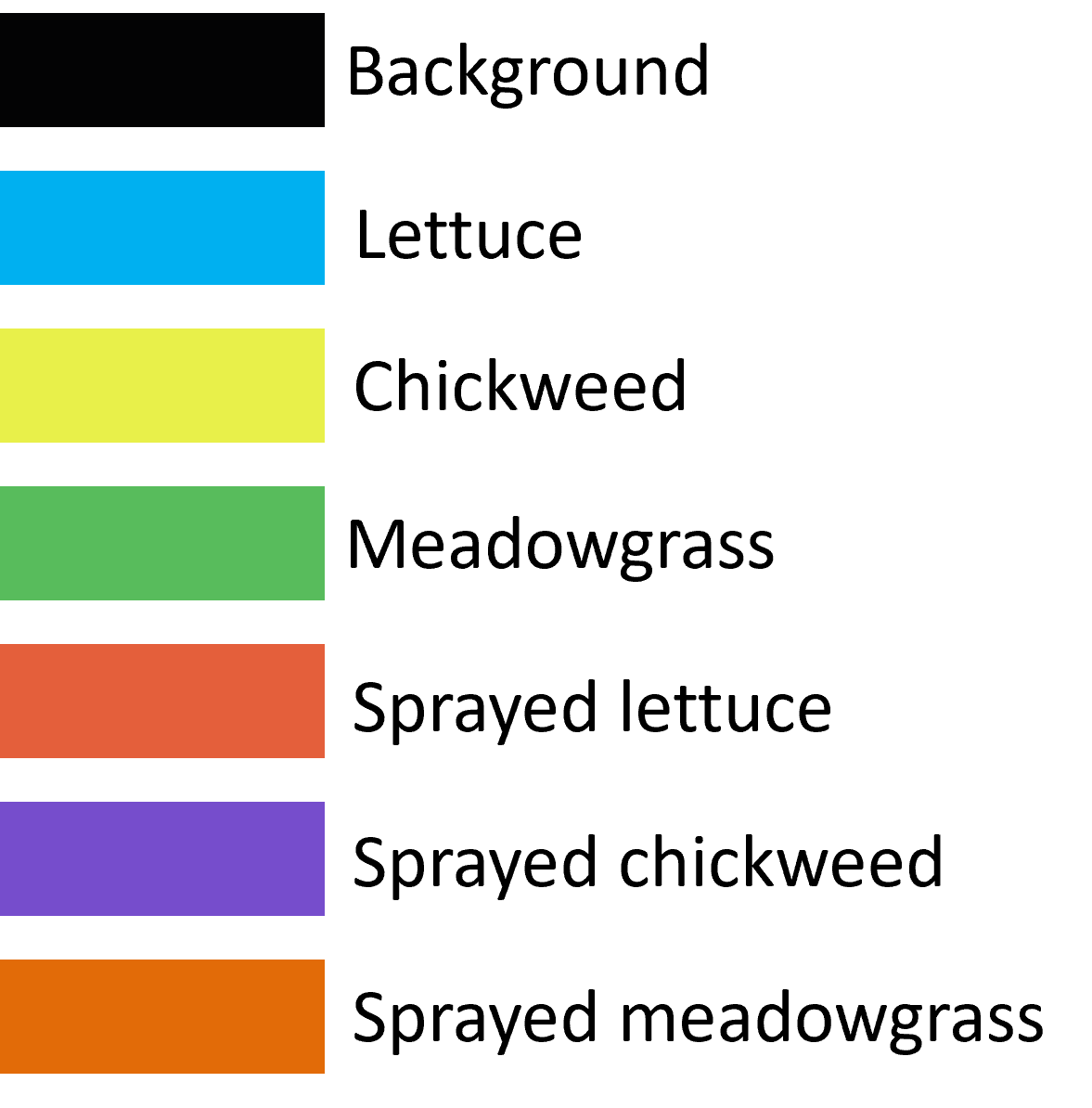}}
  
  \label{sem_seg_gt}
  \caption{RGB image \Cref{fig2a} against ground truth semantic segmentation \Cref{fig2b}.}\label{final}

\end{figure}

Using these bounding boxes objects were segmented out of the image, shown in the top right of \Cref{fig:auto}, and passed to a Segment Anything Model (SAM) \cite{kirillov2023segment} that was then able to segment each target with a higher granularity than when given the original image, an example is shown in \Cref{fig:auto} in the bottom left. When using SAM, due to the nature of the image, most of the segmentation's when not using the clusters of weeds or individual lettuces lead to unsatisfactory results, as shown in \Cref{fig:auto} in the bottom right. 
These labels were checked by the human labeler. This method and pipeline of labelling is effective as demonstrated by the ground truth shown in \Cref{final}. In the labelled image it can be seen not all chickweed or meadowgrass is sprayed despite being targeted and all lettuces are sprayed in error. 

\subsection{Deposition Estimation}
To be able to estimate deposition values the quantification of a single spray actuation needs to be validated. Therefore, quantifying a single spray actuation from the precision spraying system was completed by spraying into a container that was weighed by an analytical balance. The system sprayed 100 times into 10 identical containers. From these findings the average quantity sprayed per deposit is 20.9 $\mu$L, standard deviation is 0.16$\mu$L, standard error is 0.05$\mu$L and variance is 0.02$\mu$L thus, the average weight will be used. Visually, a single deposit on a target chickweed can be shown in \Cref{Single_deopsit}.

\begin{figure}[htbp]
    \centering
        \includegraphics[height=116px]{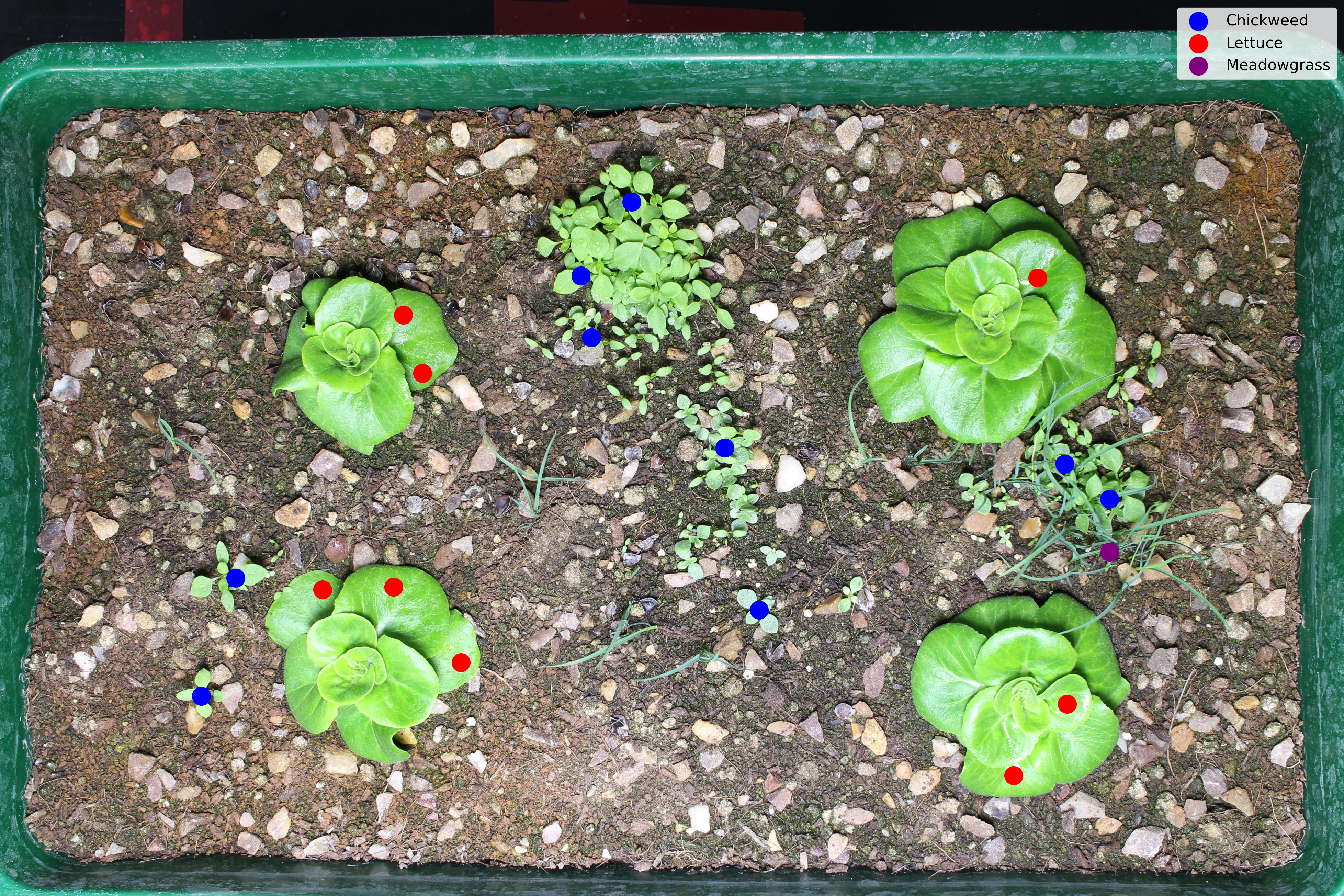}
    
    \caption{Image labelled with keypoints for Weakly Supervised Deposition Estimation  using the centre of spray actuations.}
    \label{fig:WSDE_labelled}
\end{figure}

Following this, a WSDE task has been added to the test set of the data. Key point annotations have been added. These locations are sprayed locations and are center points of spray actuations that are a minimum appropriate distance from each other considering the XY spot sprayer specifications. This task is used to be able to find an estimation of deposition values in a class-wise manner without labelling all instances in the dataset.  \Cref{fig:WSDE_labelled} shows a labelled example with a number of points on lettuces that are sprayed in error by the XY spraying system. To calculate the ground truth spraying weight for each image the number of keypoints is multiplied by the average spray actuation weight for each class. Therefore, in the example \Cref{fig:WSDE_labelled}, lettuce (in red) has been sprayed with 229.9 $\mu$L in error, chickweed (in blue) has been sprayed with 209.0 $\mu$L, and meadowgrass (in purple) has been sprayed with 20.9 $\mu$L. In total the test set contains 1212.2 $\mu$L for lettuce, 1630.2 $\mu$L for chickweed, and 188.1 $\mu$L for meadowgrass creating a total of 3030.5 $\mu$L for the test set.

\section{XAI Pipeline}\label{sec4:pipe}
To be able to interpret and complete a domain-specific WSDE task an XAI pipeline has been developed. The pipeline includes the usage of inference-only feature fusion. Within the pipeline each model and feature fusion methodology is thoroughly evaluated using segmentation and CAM metrics.

\subsection{Segmentation Architectures}
Experiments have been conducted with two semantic segmentation Deep Learning architectures, DeepLabV3 \cite{chen2017rethinking}, and Fully Convolutional Network (FCN) \cite{shelhamer2016fully}. The choice of these semantic segmentation architectures is informed from their successful deployment within agriculture \cite{LUO2023}. These architectures use differing CNN backbones, EfficientNet-B0 \cite{efficientNet}, MobileNetV3 \cite{mob}, and ResNet50 \cite{he2016deep}. Pretrained weights from PyTorch for DeepLabV3 ResNet50, MobileNetV3, and FCN ResNet50 architectures are publicly available and have been used. Using the same publicly available recipe as PyTorch we train the DeepLabV3 EfficientNet-B0, FCN EfficientNet-B0, and FCN MobileNetV3 on the Pascal class labels in the COCO dataset \cite{lin2015microsoft, Everingham2010}. These architectures are also trained with an auxiliary loss. When constructing the auxiliary loss for each architecture, that is not pretrained, it is noteworthy to mention where the auxiliary stems from. Therefore, in the EfficientNet-B0 architecture the auxiliary stem branches from stage 3 of the backbone, the MobileNetV3 and ResNet50 have the same stem points as the pretrained architectures for the pretrained DeepLabV3 and FCN at the second convolutional layer and layer 3, respectively.

\subsection{Inference-Only feature fusion}
In this paper inference-only feature fusion is explored with XAI. The semantic segmentation architectures used, utilize an auxiliary loss. Therefore, these are fused together during inference only. This means that during training, the model is optimized without the need of additional layers or computation.

Two primary fusion techniques are explored: concatenation and multiplication. Through concatenation, the auxiliary and main architecture outputs are combined along the channel dimension, allowing for the integration of information from both sources. Alternatively, multiplication involves element-wise multiplication of corresponding feature maps from the auxiliary and main outputs, facilitating a more intricate interaction between the two sets of features.

These techniques are compared not only with the segmentation metrics but also with CAM metrics. Therefore, this allows for the identification of which method is the most interpretable as well as best at the domain specific WSDE task. Essentially,  more of the architecture that is typically not used will be used. As we believe this will have a positive impact on the overall performance. This will be compared to the baseline of the main architecture output in \Cref{results}.

\subsection{Weakly Supervised Deposition Estimation}
To enable the domain-specific WSDE task, key points that are the center points of spray actuations have been labelled. Each key point labelled is an appropriate distance from each other considering the precision spraying specifications, meaning key points must be a minimum distance from each other. This task is used to be able to find an estimation of deposition values in a class-wise manner with the average spray actuation weight. 

\begin{figure}[h]
  \centering
  \subfloat[Prediction]{\includegraphics[width=0.32\linewidth]{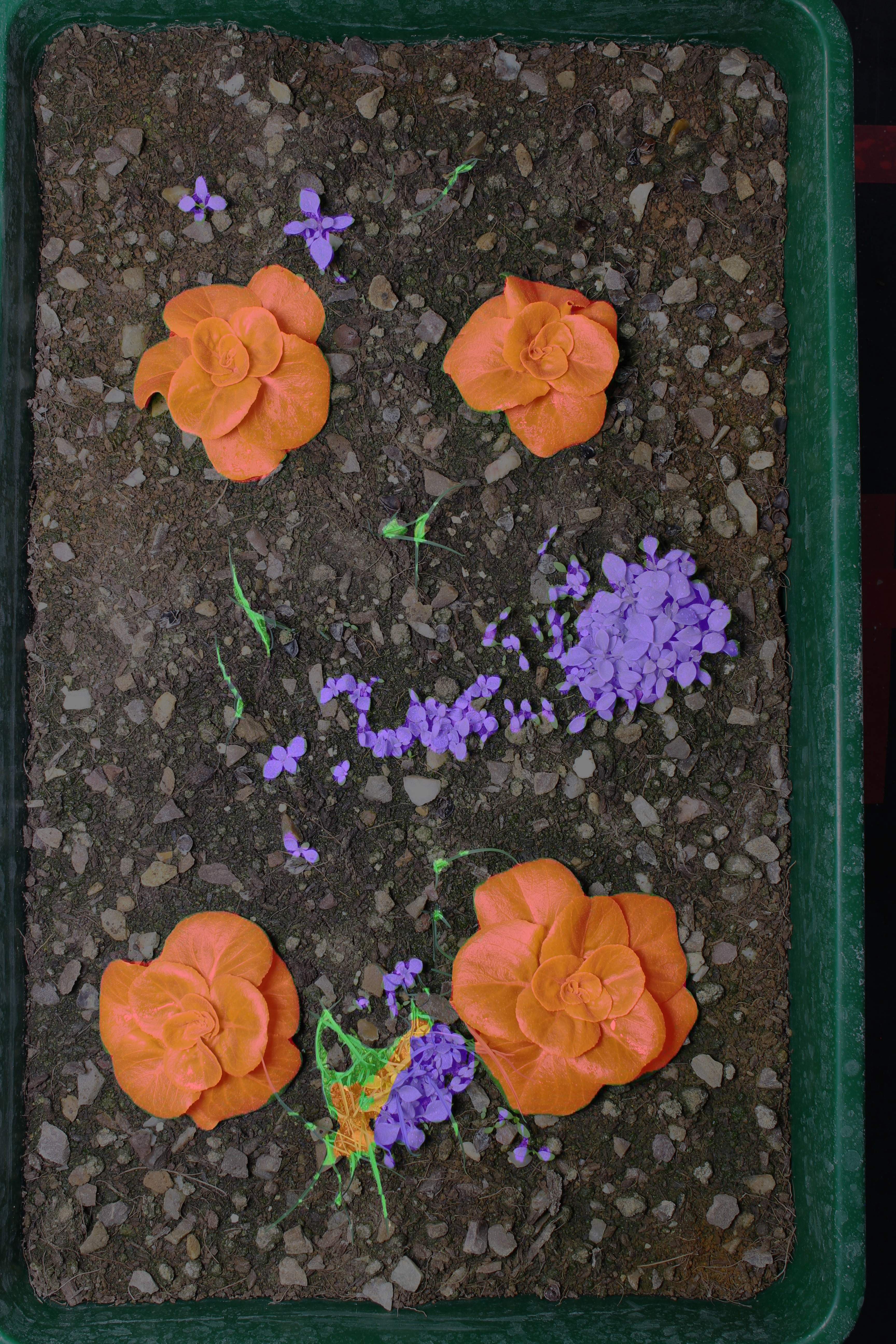}\label{pred}}
  \subfloat[CAM]{\includegraphics[width=0.32\linewidth]{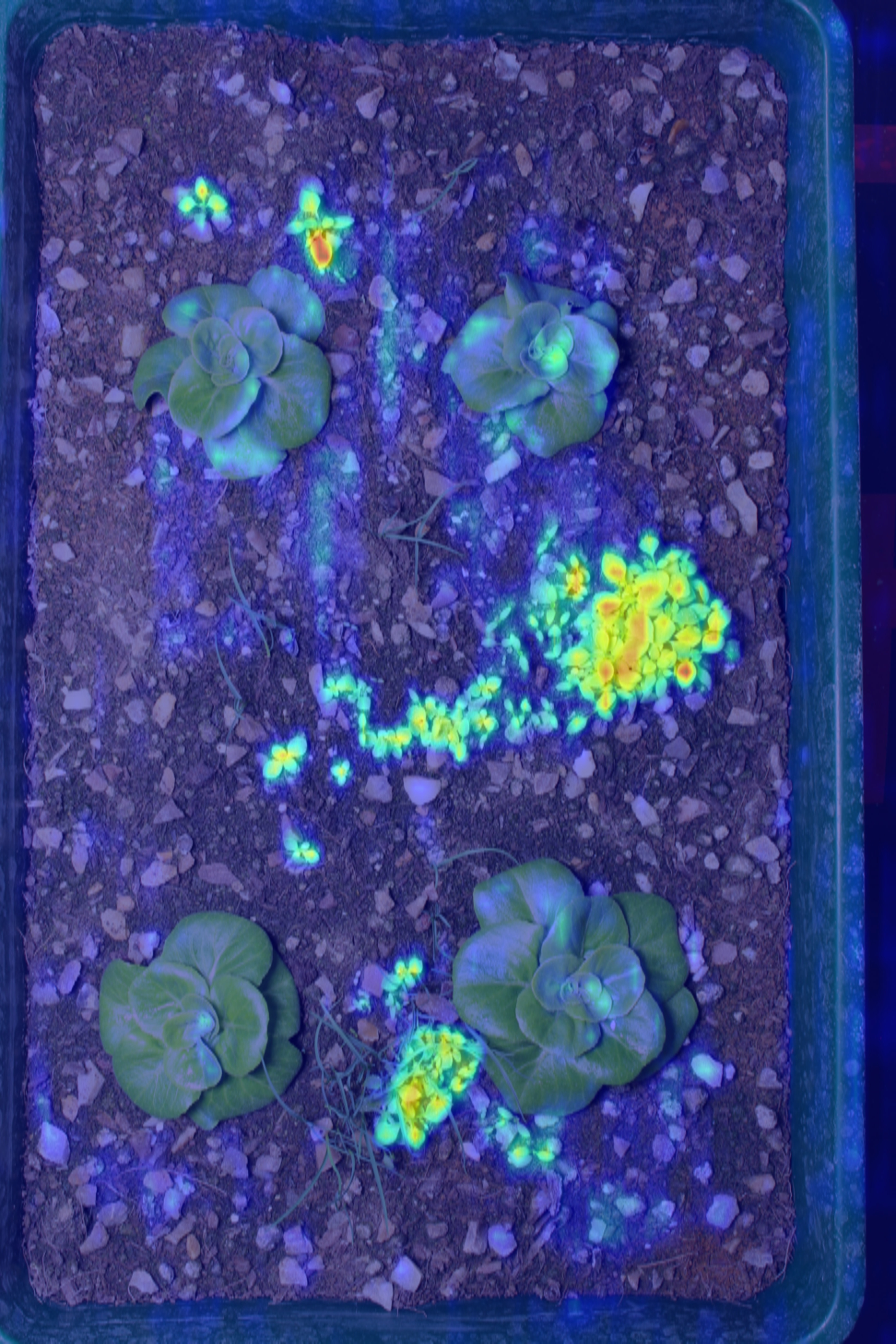}\label{cam}}
  \subfloat[Islands]{\includegraphics[width=0.32\linewidth]{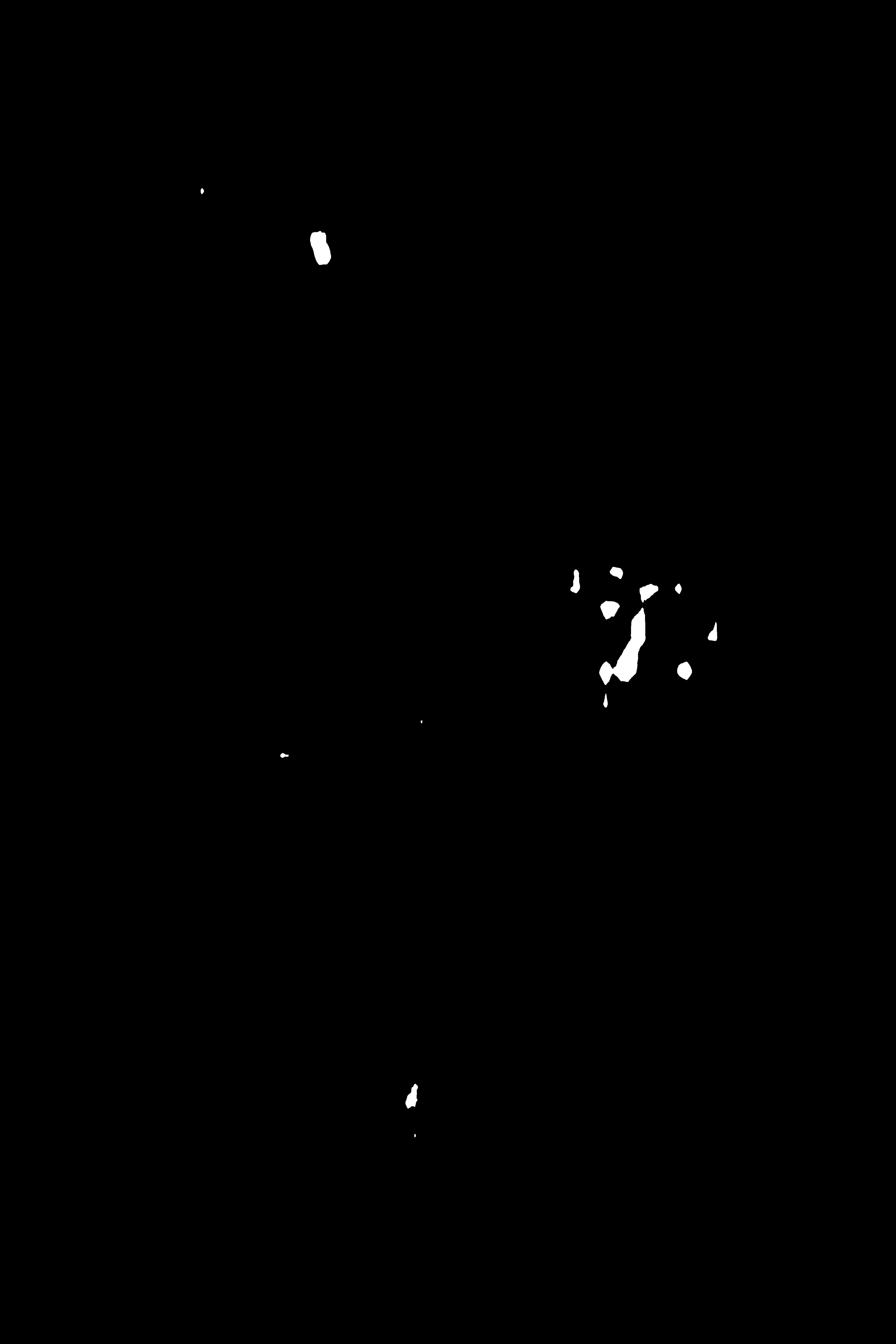}\label{islands}}
  
  \caption{Prediction from Deep Learning model \Cref{pred} with CAM for sprayed chickweed \Cref{cam} with resulting island images after thresholding and combination in \Cref{islands}.}
  \label{WSDE_EXAMPLE}
\end{figure}

To generate predictions for WSDE the Deep Learning model prediction and the top 10\% region of interest from the CAM for the specified class are multiplied. Semantic segmentation CAMs can have pixels that are not included in the desired class, therefore multiplying the CAM with the prediction allows for a better prediction. In \Cref{pred} is an example model prediction, next to it is an example of a CAM for sprayed chickweed in \Cref{cam}, finally the resulting islands from the multiplication of these is in \Cref{islands}.  After island creation, three clustering methods are compared to find the most accurate points on the resulting image. The resulting regions of interest  within the image could have several points for one cluster of weeds as shown in \Cref{islands}. As weeds were sprayed once this means clustering points that are close to each other to ensure only one prediction is made. To cluster the islands three methods are used: a baseline method using the centre points of all islands as this is not technically a clustering method, Affinity Propagation, and a thresholding methodology where the centre points of islands are removed if they are within a specified distance \cite{frey2007clustering}. The threshold distance is prior information considering the precision spraying specifications that can be altered to fit differing precision spraying systems.

WSDE evaluation is based on the original pointing game which can be used to evaluate if a CAM overlaps with objects in an image \cite{Zhang2018}. The pointing game uses an accuracy measure as shown: 

\begin{equation}
    Accuracy = \frac{\#\text{Hits}}{\#\text{Hits} + \#\text{Misses}}
\end{equation}

where a $hit$ is counted if the maximum value in a CAM lies inside of a bounding box annotation of the object class, if not a $miss$ is counted. However, for WSDE the original pointing game is not particularly useful as WSDE contains key points and not bounding boxes. Therefore, a bounding box is created for each point with  precision sprayer specifications to count hits. This will be recorded as a mean hit rate across all classes in the results.

After calculating the mean hit rate, deposition values can be calculated by multiplying the number of predictions by 20.9$\mu$L. This will then be evaluated against the ground truth weight using an absolute difference. The absolute difference is used as this illustrates the difference regardless of if the prediction is larger than the ground truth or lower as both are undesirable. The average of the absolute differences across each class for the entire test set will be reported in a class-wise manner in \Cref{results} with the best clustering methodology for each model and inference-only fusion method. This means that the best scoring models have a high mean hit rate and a low absolute difference. Using the best inference-only feature fusion method for each model there will be an  image-wise test to find the best WSDE prediction on a single tray. Using the best test prediction result, the predictions will be converted into a coverage value to show that coverage can also be created. 

\subsection{Segmentation Metrics}
\label{sem_metrics}
To evaluate semantic segmentation several different metrics are used. The metrics used are Class-wise Dice score, Mean Intersection over Union (MIoU), Pixel-wise accuracy, and Pixel Micro F1 score. Whilst there is a background class, results for it are not reported.

\begin{itemize}
    \item \textbf{Class-wise Dice Score}:
    The Class-wise Dice score measures the similarity between the predicted segmentation and the ground truth for each class individually. It calculates the overlap between the predicted and ground truth masks for each class, showing  segmentation accuracy for each specific class. This can be described as follows: 
    \begin{equation}
    Dice = \frac{{2 \times |A \cap B|}}{{|A| + |B|}}
    \end{equation}       
    Where A and B are the prediction and ground truth segmentation's, respectively.
   
    \item \textbf{Mean Intersection over Union (mIoU)}:
    mIoU computes the average IoU across all classes. IoU measures the overlap between the predicted and ground truth segmentation masks, providing an overall assessment of segmentation accuracy across all classes. IoU can be calculated as:
    \begin{align}
    IoU(A,B) = \frac{A\cap B}{A\cup B}
    \label{JACC}
    \end{align}
    Where A and B are the prediction and ground truth segmentation's, respectively. From this an average across all classes is taken. When compared to Dice score mIoU treats all errors (False Negative, or False Positive) more symmetrically whereas Dice score is more sensitive to class imbalances as True positives are weighted more heavily.
    
    \item \textbf{Pixel-wise Accuracy}:
    Pixel-wise accuracy measures the proportion of correctly classified pixels in the entire image for each class.  This metric can be computed as:
    \begin{equation}
        Pixel Accuracy = \frac{\#\text{Correctly Classified Pixels}}{\#\text{Pixels}}
    \end{equation}
    This metric is important when converting the model predictions to real world coverage as the shape of the class does not need to be considered. 
    
    \item \textbf{Micro F1 Score}:
    The micro F1 score across multiple classes is calculated by first determining True Positives (TP), False Positives (FP), and False Negatives (FN) for each pixel. TP is the number of pixels classified correctly by the model; FP is counted when the model incorrectly predicts a pixel as belonging to a certain class, but that pixel actually belongs to a different class, FN is counted when the model fails to correctly predict a pixel as belonging to a certain class, instead predicting it as belonging to a different class. Using these, precision ($P$) and recall ($R$) are computed for each class using the formulas: 
    \begin{equation}
        P = \frac{TP}{TP + FP}
    \end{equation}
    \begin{equation}
        R = \frac{TP}{TP + FN}
    \end{equation}
    Subsequently, the F1 score for each class is obtained using the formula:
    \begin{equation}
        F1 = 2 * \frac{(P*R)}{(P+R)}
    \end{equation}
    To calculate the micro F1 score, we aggregate the TP, FP, and FN scores across all classes before computing precision and recall. This means summing TP, FP, and FN for all classes and then using these aggregated counts to compute a single precision and recall value:
    
    \begin{equation}
        P_{micro} = \frac{\sum TP}{\sum TP + \sum FP}
    \end{equation}
    \begin{equation}
        R_{micro} = \frac{\sum TP}{\sum TP + \sum FN}
    \end{equation}
    
    The micro F1 score is then calculated using these micro-averaged precision and recall values:
    
    \begin{equation}
        F1_{micro} = 2 * \frac{(P_{micro} * R_{micro})}{(P_{micro} + R_{micro})}
    \end{equation}
    
    Micro F1 is selected as the overall performance of models is imperative.

\end{itemize}

\subsection{CAM Metrics}
\label{cam_metrics}
CAMs are evaluated using Deletion, Insertion, and the domain-specific WSDE task. 

For Deletion and Insertion, the confidence from a Deep Learning model prediction is recorded. Using MoRF areas from a CAM are  removed or inserted increasing the area by 1\% for both until the entire image is deleted or inserted. After plotting the confidence values from the Deep Learning model against the amount of each image that is deleted or inserted, the Area Under the Curve (AUC) is calculated using the Trapezoidal Rule as follows:

\begin{align}
AUC = \frac{h}{2} \left[y_0 + 2 \left(y_1 + y_2 + y_3 + \cdots + y_{n-1}\right) + y_n\right]
\label{area}
\end{align}
where y is the prediction confidence, n is equal to the number of plotted points, and h is equal to the increase in deletion or insertion change. Therefore, Deletion scores that are lower are better and Insertion scores that are higher are better. An average across all classes for each is taken and reported in \Cref{results}.

\begin{table*}
\centering
\caption{Class-wise Pixel Accuracy and micro F1 Scores (bold indicates performance better than baseline)}
\label{tab:pixel_f1}
\begin{tabular}{llrrrrrrr}
Model & Fusion & Lettuce & Chickweed & Meadowgrass & Sprayed Lettuce & Sprayed Chickweed & Sprayed Meadowgrass & Micro F1 \\ \hline
\multirow{4}{*}{\begin{tabular}[c]{@{}l@{}}DeepLabv3 \\ (EfficientNetB0)\end{tabular}} & Out & 96.78 & 83.86 & 46.32 & 98.62 & 85.63 & 69.16 & 97.44 \\
 & Aux & 52.68 & 48.64 & \textbf{64.04} & 58.21 & 38.39 & 0.00 & 95.08 \\
 & Add & \textbf{97.40} & \textbf{86.73} & \textbf{55.24} & \textbf{98.77} & \textbf{88.02} & 1.30 & \textbf{98.81} \\
 & Multi & 94.84 & 82.63 & \textbf{51.09} & 97.50 & 84.94 & 0.02 & \textbf{98.46} \\ \cline{2-9} 
\multirow{4}{*}{\begin{tabular}[c]{@{}l@{}}DeepLabv3 \\ (MobileNetV3)\end{tabular}} & Out & 97.19 & 83.80 & 42.99 & 84.95 & 88.24 & 65.80 & 98.23 \\
 & Aux & 96.10 & 71.49 & \textbf{61.17} & 0.02 & 22.88 & 0.00 & 95.13 \\
 & Add & \textbf{97.44} & \textbf{88.15} & \textbf{52.30} & 83.95 & \textbf{89.84} & 1.51 & \textbf{98.47} \\
 & Multi & 95.09 & 78.98 & \textbf{48.21} & 82.66 & 84.43 & 0.05 & 98.04 \\ \cline{2-9} 
\multirow{4}{*}{\begin{tabular}[c]{@{}l@{}}DeepLabv3 \\ (ResNet50)\end{tabular}} & Out & 92.98 & 92.18 & 61.84 & 93.59 & 90.58 & 71.93 & 98.73 \\
 & Aux & 85.42 & 44.81 & 60.89 & 81.75 & 70.97 & 0.00 & 97.57 \\
 & Add & \textbf{93.49} & 89.46 & 61.45 & 90.70 & 88.77 & 32.33 & 98.70 \\
 & Multi & 92.67 & 87.42 & 59.58 & 90.95 & 87.69 & 41.96 & 98.60 \\ \hline
\multirow{4}{*}{\begin{tabular}[c]{@{}l@{}}FCN \\ (EfficientNetB0)\end{tabular}} & Out & 94.51 & 74.00 & 35.94 & 91.02 & 71.90 & 59.96 & 97.91 \\
 & Aux & \textbf{97.48} & 45.15 & \textbf{63.57} & 0.14 & 49.45 & 0.00 & 95.31 \\
 & Add & \textbf{96.39} & \textbf{83.11} & \textbf{46.05} & \textbf{91.96} & \textbf{80.14} & 5.57 & \textbf{98.61} \\
 & Multi & 93.08 & \textbf{79.85} & \textbf{43.91} & 89.89 & \textbf{77.11} & 0.69 & \textbf{98.23} \\ \cline{2-9} 
\multirow{4}{*}{\begin{tabular}[c]{@{}l@{}}FCN \\ (MobileNetV3)\end{tabular}} & Out & 95.01 & 74.67 & 35.98 & 93.42 & 77.30 & 69.46 & 98.16 \\
 & Aux & 85.90 & 70.57 & \textbf{64.08} & 16.60 & 27.61 & 0.00 & 95.36 \\
 & Add & \textbf{96.75} & \textbf{85.23} & \textbf{44.03} & \textbf{93.96} & \textbf{85.96} & 4.39 & 94.80 \\
 & Multi & 93.53 & 73.75 & \textbf{47.74} & 92.21 & \textbf{81.60} & 0.40 & \textbf{98.40} \\ \cline{2-9} 
\multirow{4}{*}{\begin{tabular}[c]{@{}l@{}}FCN \\ (ResNet50)\end{tabular}} & Out & 85.97 & 84.89 & 62.11 & 88.75 & 68.41 & 39.54 & 98.01 \\
 & Aux & 82.39 & 83.78 & \textbf{66.06} & 81.91 & 57.42 & 0.00 & 97.54 \\
 & Add & 85.56 & \textbf{86.57} & \textbf{64.45} & 86.98 & 65.56 & 2.10 & 97.92 \\
 & Multi & 84.74 & \textbf{85.13} & \textbf{62.87} & 87.27 & 65.73 & 1.74 & 97.84
\end{tabular}
\end{table*}

\section{Results} 
\label{results}
When comparing the outcomes of our experiments, we have designated the feature fusion method for each model as follows: OUT is the main traditional output (the baseline), AUX is the auxiliary output, ADD is the concatenation of the output and auxiliary, and MULTI indicates the multiplication of both the output and auxiliary. These designations are capitalized for clarity. The results are split into semantic segmentation scores, inference-only feature fusion interpretability, and WSDE.

\subsection{Semantic Segmentation}
\Cref{tab:pixel_f1} presents the class-wise pixel accuracies and micro F1 scores, \Cref{tab:dice} displays the class-wise Dice scores and mIoU scores. Notably, feature fusion scores that surpass the baseline are highlighted in bold for enhanced clarity for both tables.

 \begin{table*}
\centering
\caption{Class-wise Dice and mIoU Scores (bold indicates performance better than baseline)}
\label{tab:dice}
\begin{tabular}{llrrrrrrr} 
Model & Fusion & Lettuce & Chickweed & Meadowgrass & Sprayed Lettuce & Sprayed Chickweed & Sprayed Meadowgrass & mIoU \\ \hline
\multirow{4}{*}{\begin{tabular}[c]{@{}l@{}}DeepLabv3 \\ (EfficientNetB0)\end{tabular}} & Out & 93.0 & 43.4 & 31.0 & 91.6 & 63.3 & 56.6 & 64.9 \\
 & Aux & 56.5 & 32.8 & \textbf{37.6} & 52.2 & 28.8 & 0.0 & 37.8 \\
 & Add & \textbf{96.2} & \textbf{50.3} & \textbf{36.5} & \textbf{96.4} & \textbf{73.1} & 2.5 & \textbf{67.0} \\
 & Multi & 92.3 & \textbf{51.6} & \textbf{40.6} & \textbf{92.5} & \textbf{73.5} & 0.0 & \textbf{65.9} \\ \cline{2-9} 
\multirow{4}{*}{\begin{tabular}[c]{@{}l@{}}DeepLabv3 \\ (MobileNetV3)\end{tabular}} & Out & 91.4 & 48.7 & 33.1 & 89.3 & 68.8 & 65.4 & 68.9 \\
 & Aux & 70.8 & 34.5 & \textbf{36.6} & 0.0 & 24.6 & 0.0 & 34.3 \\
 & Add & \textbf{92.1} & \textbf{50.4} & \textbf{36.0} & \textbf{89.8} & \textbf{73.0} & 3.0 & 64.3 \\
 & Multi & 88.6 & \textbf{51.0} & \textbf{36.7} & 82.2 & \textbf{73.8} & 0.0 & 61.8 \\ \cline{2-9} 
\multirow{4}{*}{\begin{tabular}[c]{@{}l@{}}DeepLabv3 \\ (ResNet50)\end{tabular}} & Out & 92.8 & 54.6 & 44.0 & 91.5 & 67.5 & 64.8 & 76.7 \\
 & Aux & 85.2 & 39.9 & 38.4 & 80.9 & 40.6 & 0.0 & 51.5 \\
 & Add & 92.3 & \textbf{55.3} & 40.6 & 90.7 & 67.1 & 47.1 & 71.1 \\
 & Multi & 89.6 & \textbf{55.5} & \textbf{44.2} & 89.9 & \textbf{67.7} & 33.5 & 69.6 \\ \hline
\multirow{4}{*}{\begin{tabular}[c]{@{}l@{}}FCN \\ (EfficientNetB0)\end{tabular}} & Out & 92.2 & 45.0 & 29.3 & 91.7 & 59.2 & 62.8 & 65.1 \\
 & Aux & 71.4 & 31.4 & \textbf{38.3} & 0.3 & 34.3 & 0.0 & 36 \\
 & Add & \textbf{94.1} & \textbf{50.6} & \textbf{34.2} & \textbf{93.6} & \textbf{66.5} & 10.3 & 64.5 \\
 & Multi & 89.8 & \textbf{51.7} & \textbf{39.9} & 89.3 & \textbf{66.6} & 0.2 & 62.8 \\ \cline{2-9} 
\multirow{4}{*}{\begin{tabular}[c]{@{}l@{}}FCN \\ (MobileNetV3)\end{tabular}} & Out & 93.7 & 45.7 & 31.6 & 93.4 & 63.5 & 67.6 & 68.7 \\
 & Aux & 69.0 & 37.6 & \textbf{38.8} & 25.0 & 27.4 & 0.0 & 37.5 \\
 & Add & \textbf{95.2} & \textbf{51.6} & \textbf{34.4} & \textbf{95.1} & \textbf{70.6} & 8.3 & 66.3 \\
 & Multi & 91.9 & \textbf{49.9} & \textbf{38.2} & 91.8 & \textbf{69.4} & 0.1 & 63.6 \\ \cline{2-9} 
\multirow{4}{*}{\begin{tabular}[c]{@{}l@{}}FCN \\ (ResNet50)\end{tabular}} & Out & 87.6 & 46.9 & 40.8 & 85.2 & 54.3 & 51.1 & 63.3 \\
 & Aux & 83.4 & 43.6 & 40.2 & 79.3 & 50.5 & 0.0 & 53.7 \\
 & Add & 86.8 & 46.7 & 40.0 & 84.0 & \textbf{54.5} & 4.1 & 57.5 \\
 & Multi & 85.4 & 47.2 & \textbf{41.8} & 82.9 & \textbf{55.4} & 1.5 & 57.3
\end{tabular}
\end{table*}

A significant observation from the results concerns the AUX feature fusions inability to predict the final class, sprayed meadowgrass, across all architectures tested, with all evaluation metrics. This suggests that the layers between the auxiliary output and traditional output are crucial in the learning process of the final class across both architectures tested with the EfficientNet-B0, MobileNetV3, and ResNet50 backbones.

The DeebLabV3 (EfficientNet-B0) with the ADD feature fusion achieved the highest micro F1 score across all models and backbones at 98.81\%. When comparing with other feature fusion methods within the same backbone and architecture, the ADD fusion outperforms all other fusion methods across most classes (except sprayed meadowgrass) in terms of pixel accuracies. Furthermore, the ADD and MULTI feature fusion exhibit higher mIoU scores compared to the baseline, at 67.0\%, 65.9\%, and 64.9\%, respectively. Moreover, both ADD and MULTI feature fusion methods yield higher dice scores across most classes, with exceptions for lettuce in the case of MULTI and sprayed meadowgrass for ADD and MULTI.

For the DeepLabV3 (MobileNetV3), the ADD feature fusion demonstrates better pixel accuracies across most classes compared to other fusion methods and the baseline in \Cref{tab:pixel_f1}. The improvement in micro F1 scores is marginal with the ADD feature fusion compared to the baseline, ADD scores 98.47\% and the baseline scores 98.23\%. There is also improvement in the Dice scores for the ADD feature fusion across the majority of classes when compared to the baseline.

In the case of the DeepLabV3 (ResNet50), the baseline achieves high accuracy, with no inference-only feature fusion method showing improvement in terms of segmentation metrics. This is also reflected in the mIoU score, which reaches 76.7\%, the highest among all architectures and backbones tested.

The FCN (EfficientNet-B0) architecture, using the ADD feature fusion shows improvements in pixel accuracies, Dice scores, and micro F1 scores compared to the baseline. The largest increase in pixel accuracy is with meadowgrass with a 10.11\% increase. Micro F1 increases from 97.91\% to 98.61\%, and Dice for sprayed chickweed increases by 7.3\% when using the ADD feature fusion. Similarly, the MULTI feature fusion increases in terms of pixel accuracies and micro F1 score. Micro F1 increases from 97.91\% to 98.23\% and the pixel accuracy for meadowgrass increases by 9.7\% when using the MULTI feature fusion.

The FCN (MobileNetV3) improves from the baseline with the ADD feature fusion when considering pixel accuracies and Dice scores. Pixel accuracy increases by 10.56\% for chickweed, and Dice increases by 7.1\% for sprayed chickweed with the ADD feature fusion. The MULTI feature fusion also improves from the baseline, with micro F1 scores from 98.16\% to 98.40\%.

Similar to DeepLabV3 (ResNet50), the FCN (ResNet50) baseline demonstrates high accuracy, with no notable improvements with inference-only feature fusion methods. These results suggest that with the ResNet50 backbone, the best outcomes are achieved when feature fusion is not employed, possibly due to the ResNet50 backbone being larger when compared to other tested backbones.

\subsection{Inference-only feature fusion interpretability}

\begin{figure*}[th]
  \centering
  \subfloat[AblationCAM scores.]{\includegraphics[width=0.5\linewidth]{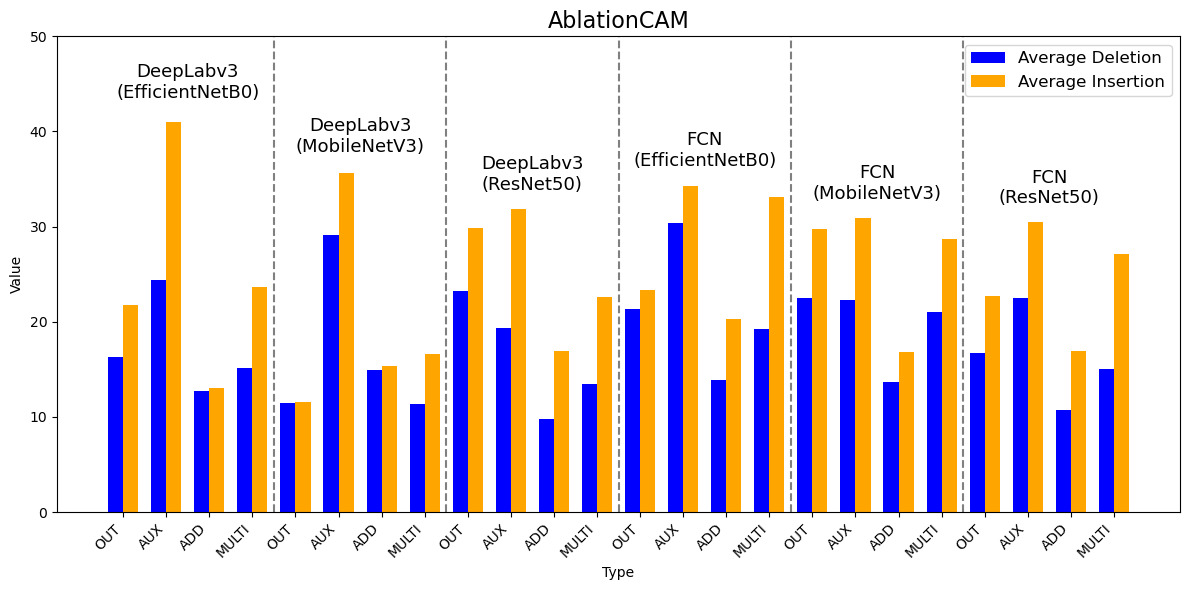}\label{fig7a}}
  \subfloat[ScoreCAM scores.]{\includegraphics[width=0.5\linewidth]{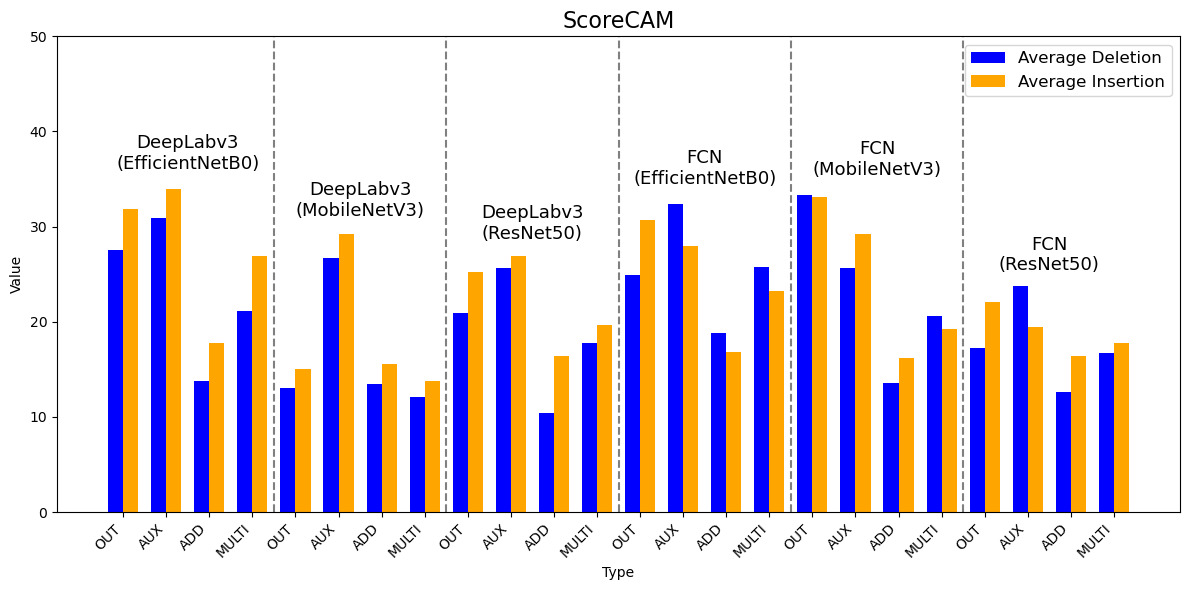}\label{fig7b}}
  \caption{AblationCAM average Deletion and Insertion \Cref{fig7a} against ScoreCAM average Deletion and Insertion \Cref{fig7b}.}
  \label{del_ins_scores}
\end{figure*}

\Cref{del_ins_scores} shows that AblationCAM generally produces more interpretable CAMs than ScoreCAM. It can be seen that all models for AblationCAM are interpretable, as Deletion is lower than Insertion, except for the DeepLabV3 (MobileNetV3) using the baseline. ScoreCAM has six models that are not interpretable. To compare CAM methods effectively the difference between Deletion and Insertion is used, larger differences are better.

When considering the DeepLabV3 (EfficientNet-B0), using AblationCAM, the easiest feature fusion to interpret is the AUX which has a difference of 16.6\%, this is the highest across all models tested. The next easiest to interpret is the MULTI fusion with a difference of 8.5\%. Next the baseline which scores a difference of 5.5\%, then finally ADD scores a difference of 0.2\%. When using ScoreCAM, the fusion differences are 5.9\% for MULTI, 4.3\% for the baseline, 4.0\% for ADD, and 3.0\% for AUX. This means with the DeepLabV3 (EfficientNet-B0) it is easier to interpret the CAMs with AblationCAM as the differences for the baseline, AUX, and MULTI are larger.

Moving to the DeepLabV3(MobileNetV3) with AblationCAM the only model that is not interpretable appears when using the baseline. This means that models are easier to interpret when using inference-only feature fusion as the AUX scores a difference of 6.5\%, MULTI scores a difference of 5.2\%, and ADD scores 0.4\%. Whereas the baseline has the same score for Deletion and Insertion thus not interpretable. For ScoreCAM, however, each method is interpretable with AUX scoring a difference of 2.5\%, ADD scoring 2.1\%, the baseline scoring 2.0\%, and MULTI scoring 1.7\%. 

The DeepLabV3 (ResNet50) is interpretable with both AblationCAM and ScoreCAM as all fusion methods have a lower average Deletion than Insertion. The score differences, when considering AblationCAM, are 12.6\% for AUX, 9.1\% for MULTI, 7.1\% for ADD, and 6.7\% for the baseline. ScoreCAM has differences of 6.0\% for the ADD fusion, 4.3\% for the baseline, 1.8\% for MULTI, and 1.3\% for AUX. This would suggest that as the differences are much larger for AblationCAM when compared to ScoreCAM for all backbones in the DeepLabV3 architecture AblationCAM is better at generating representative CAMs for semantic segmentation.

Similarly, the FCN (EfficientNet-B0) with AblationCAM has interpretable CAMs where the difference between Deletion and Insertion for fusion methods is better than ScoreCAM methods. Specifically, with AblationCAM, the FCN (EfficientNet-B0) with the MULTI fusion has a difference of 13.9\%, ADD has a difference of 6.5\%, AUX has a difference of 3.9\%, and the baseline has a difference of 2.0\%. This means that all feature fusion methods are more interpretable than the baseline with AblationCAM. However, when considering ScoreCAM, the baseline has a difference of 5.7\%, and the other tested fusion methods are not interpretable.

The FCN (MobileNetV3) with AblationCAM has all fusion methods as interpretable. Notably, the MULTI fusion scores a difference of 13.9\%, ADD scores 6.5\%, AUX scores 3.9\%, and the baseline has a difference of 2.0\%. Again, showing that AblationCAM is more interpretable with the proposed inference-only feature fusion. Furthermore, not all methods are interpretable with ScoreCAM, AUX scores a difference of 3.5\%, ADD scores 2.6\%, and the baseline and MULTI are not interpretable. 

Finally, the FCN (ResNet50) with AblationCAM has interpretable CAMs for all methods. In particular, the MULTI scores a difference of 12.1\%, AUX has a difference of 8.0\%, ADD scores 6.2\%, and the baseline scores 6.0\%. ScoreCAM CAMs are not all interpretable but the baseline scores a difference of 4.8\%, ADD has a score of 3.8\%, and MULTI has a score of 1.0\%. Thus, it can be concluded that for the FCN architecture feature fusion and AblationCAM has a positive impact where CAMs are more interpretable compared to the baseline.

As AblationCAM is easier to interpret using Deletion and Insertion in comparison to ScoreCAM, AblationCAM will be exclusively used in the WSDE task.

\begin{table*}[b]
\caption{AblationCAM Average Deposition Estimation (bold indicates best score)}
\centering
\label{tab:AblationDIFF}
\begin{tabular}{lllrrrrr}
Model & Fusion & Method & Lettuce & Chickweed & Meadowgrass & Total & Mean Hit Rate \\ \hline
\multirow{4}{*}{\begin{tabular}[c]{@{}l@{}}DeepLabv3 \\ (EfficientNetB0)\end{tabular}} & Out & Centres & 151.5 & 141.5 & 23.5 & 316.1 & 11.6 \\
 & Aux & Affinity & 97.0 & \textbf{50.3} & 18.4 & \textbf{164.6} & 34.4 \\
 & Add & Centres & \textbf{39.8} & 110.3 & 23.5 & 172.4 & \textbf{36.1} \\
 & Multi & Centres & 115.4 & 170.5 & \textbf{18.3} & 303.1 & 34.5 \\ \cline{2-8} 
\multirow{4}{*}{\begin{tabular}[c]{@{}l@{}}DeepLabv3 \\ (MobileNetV3)\end{tabular}} & Out & Centres & 130.9 & 68.8 & 172.4 & 371.0 & \textbf{33.8} \\
 & Aux & Centres & 768.6 & 421.1 & \textbf{23.5} & 1,212.2 & 29.2 \\
 & Add & Centres & \textbf{94.2} & \textbf{66.0} & \textbf{23.5} & \textbf{182.9} & 25.6 \\
 & Multi & Centres & 133.7 & 222.6 & \textbf{23.5} & 378.8 & 32.0 \\ \cline{2-8} 
\multirow{4}{*}{\begin{tabular}[c]{@{}l@{}}DeepLabv3 \\ (ResNet50)\end{tabular}} & Out & Centres & 123.1 & 144.4 & 23.5 & 290.0 & 30.8 \\
 & Aux & Centres & \textbf{52.6} & 146.8 & \textbf{18.3} & 216.8 & 26.9 \\
 & Add & Centres & 68.3 & 123.5 & 23.7 & \textbf{214.2} & \textbf{37.4} \\
 & Multi & Affinity & 204.3 & \textbf{100.0} & 23.6 & 326.6 & 36.8 \\ \hline
\multirow{4}{*}{\begin{tabular}[c]{@{}l@{}}FCN \\ (EfficientNetB0)\end{tabular}} & Out & Centres & \textbf{63.3} & 162.2 & 18.4 & 243.0 & 26.9 \\
 & Aux & Centres & 76.1 & 152.0 & 23.5 & 250.8 & 25.5 \\
 & Add & Affinity & 71.0 & \textbf{66.0} & 21.0 & \textbf{156.8} & \textbf{38.3} \\
 & Multi & Centres & 118.1 & 84.1 & \textbf{18.3} & 219.5 & 28.9 \\ \cline{2-8} 
\multirow{4}{*}{\begin{tabular}[c]{@{}l@{}}FCN \\ (MobileNetV3)\end{tabular}} & Out & Affinity & 151.5 & 423.5 & 23.5 & 598.3 & 6.9 \\
 & Aux & Centres & 70.9 & 126.2 & 96.7 & \textbf{292.6} & \textbf{34.4} \\
 & Add & Centres & 151.5 & 211.7 & 23.5 & 386.7 & 3.1 \\
 & Multi & Centres & 65.7 & 126.1 & 240.4 & 431.1 & 30.5 \\ \cline{2-8} 
\multirow{4}{*}{\begin{tabular}[c]{@{}l@{}}FCN \\ (ResNet50)\end{tabular}} & Out & Affinity & 102.2 & 123.3 & \textbf{20.9} & \textbf{245.6} & 25.5 \\
 & Aux & Affinity & 151.5 & 175.3 & 23.5 & 350.1 & 6.9 \\
 & Add & Centres & \textbf{94.4} & \textbf{118.3} & 81.0 & 292.6 & \textbf{32.6} \\
 & Multi & Centres & 157.0 & 287.5 & 23.5 & 467.6 & 11.0

\end{tabular}
\end{table*}

\subsection{Weakly Supervised Deposition Estimation}
In \Cref{tab:AblationDIFF} the AblationCAM average $\mu$L absolute differences and mean hit rate are reported. The best results are in bold. 

As shown in \Cref{tab:AblationDIFF} the best performing model for mean hit rate and total average absolute difference is the FCN (EfficientNet-B0) with the ADD feature fusion using Affinity propagation. The total average absolute difference is 156.8 $\mu$L and the mean hit rate is 38.3\%. When looking at class-specific average absolute differences the ADD fusion scores 71.0 $\mu$L for lettuce, 66.0 $\mu$L for chickweed, and 20.9 $\mu$L for meadowgrass. The best score for the backbone and model when considering the lettuce class is the baseline scoring 63.3 $\mu$L, and the lowest for the meadowgrass class is from the MULTI fusion with a score of 18.3 $\mu$L.

The DeepLabV3 (EfficientNet-B0) the best total absolute difference is with the AUX feature fusion at 164.6 $\mu$L using Affinity propagation. When looking at class-specific average absolute differences the model scores 39.8  $\mu$L for lettuce, 110.3 $\mu$L for chickweed, and 18.3 $\mu$L for meadowgrass. The lowest score for the backbone and model when considering chickweed is the AUX feature fusion scoring 50.3 $\mu$L. However, the best mean hit rate is with the ADD feature fusion at 36.1\% using the center points. This shows the deposition estimation from the AUX fusion method is closer but it is not as accurate in its spatial location when compared to the ADD feature fusion. 

The DeepLabV3 (MobileNetV3) has the best total absolute difference with the ADD feature fusion using centers at 182.9 $\mu$L. When considering the lettuce the average absolute difference with the ADD fusion is 94.2 $\mu$L, 66.0 $\mu$L for the chickweed, and 23.5 $\mu$L for meadowgrass. However, the best mean hit rate is from the baseline with the center points at 33.8\%. 

The DeepLabV3 (ResNet50) performs best with the ADD feature fusion using the center points at 214.2 $\mu$L, and 37.4\% for the total absolute difference, and mean hit rate, respectively. Considering class-specific absolute difference scores, the AUX fusion scores best with 52.6 $\mu$L, the MULTI scores best for chickweed at 100.0 $\mu$L, and the AUX scores best with 18.3 $\mu$L for meadowgrass.

With the FCN (MobileNetV3) the best results are with the AUX feature fusion for both total absolute difference and mean hit rate at 292.6 $\mu$L and 34.4\%, respectively using the center points. However, in a class-specific absolute difference comparison none of the AUX results are best. The best lettuce absolute difference is 65.7 $\mu$L with the MULTI, for chickweed  MULTI scores best at 126.1 $\mu$L, and the ADD scores best for meadowgrass with 23.5 $\mu$L.

Finally, the FCN (ResNet50) performs best with a total absolute difference of 245.6 $\mu$L with the baseline using Affinity propagation. However, the baseline does not perform best in all classes. It only does for meadowgrass with a difference of 20.9 $\mu$L. The best performing for the lettuce and chickweed class is the ADD fusion with 94.4 $\mu$L, and 118.3 $\mu$L, respectively. The best mean hit rate for the model uses the ADD feature and the center points scoring 32.6\%.

\begin{figure*}[th]
  \centering
  \subfloat[Model prediction.]{\includegraphics[height=0.4\linewidth]{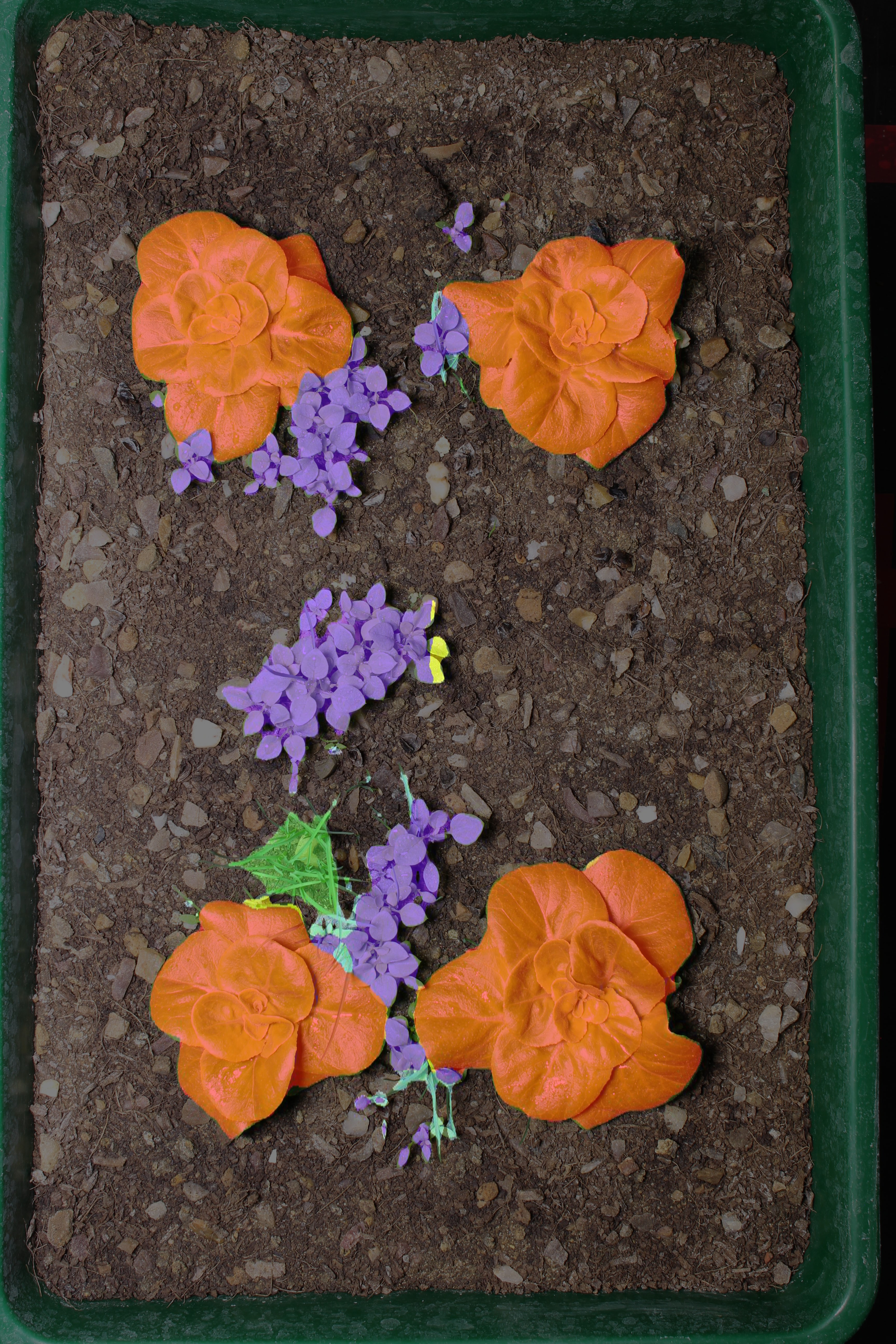}\label{fig3a}}
  \subfloat[Sprayed Chickweed CAM.]{\includegraphics[height=0.4\linewidth]{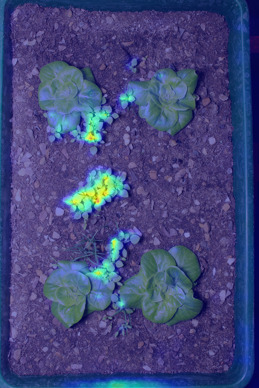}\label{fig3b}}
   \subfloat[Sprayed Lettuce CAM.]{\includegraphics[height=0.4\linewidth]{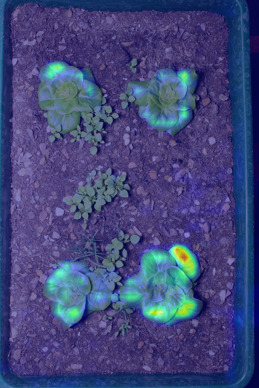}\label{fig3c}}
  \subfloat[Deposition prediction and Ground Truth.]{\includegraphics[height=0.4\linewidth]{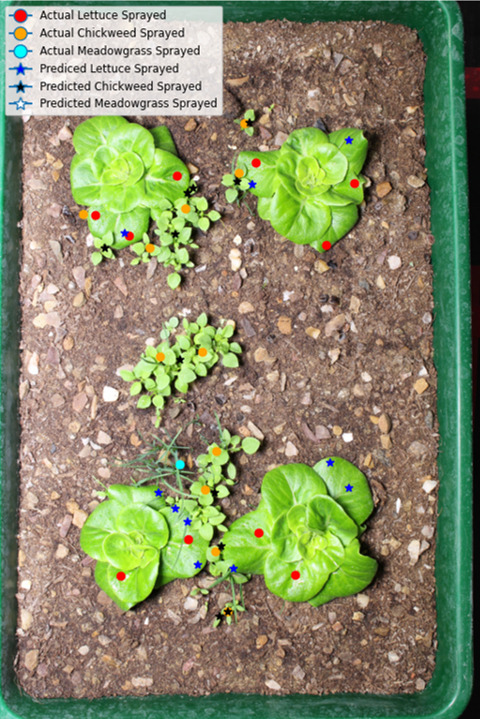}}\label{fig3d}
  \caption{Images generated to get to deposition values related to \Cref{tray_coverage}.}
  \label{coverage_depo_stuff}
\end{figure*}

\begin{table}
\centering
\caption{Image-wise Absolute Difference}
\label{tab:abs_single}
\begin{tabular}{lllrrr} 
Model & Fusion & Method & Tray & \begin{tabular}[r]{@{}r@{}}Absolute\\ Difference\end{tabular} & \begin{tabular}[r]{@{}r@{}}Mean \\Hit Rate\end{tabular} \\ 
\hline
\begin{tabular}[l]{@{}l@{}}FCN \\(EfficientNetB0)\end{tabular} & Aux & Centres & 3 & 62.7 & 66.6\% \\ 

\begin{tabular}[l]{@{}l@{}}FCN\\(ResNet50)\end{tabular} & Multi & Affinity & 5 & 41.8 & 36.1\% \\ 

\begin{tabular}[l]{@{}l@{}}DeeplabV3 \\(MobileNetV3)\end{tabular} & Aux & Centres & 7 & 83.6 & 46.3\%\\ 

\begin{tabular}[l]{@{}l@{}}DeeplabV3\\(EfficientNetB0)\end{tabular} & Multi & Centres & 16 & 41.8  & 62.5\%\\ 

\begin{tabular}[l]{@{}l@{}}FCN\\(ResNet50)\end{tabular} & Multi & Affinity & 27 & 20.9 & 46.9\% \\ 

\begin{tabular}[l]{@{}l@{}}DeeplabV3\\(EfficientNetB0)\end{tabular} & Add & Affinity & 43 & 41.8 & 40.4\%\\ 

\begin{tabular}[l]{@{}l@{}}FCN\\(ResNet50)\end{tabular} & Add & Affinity & 123 & 41.8 &50.0 \%\\

\end{tabular}
\end{table}

Following this, an exploration into the best absolute difference using image-specific tests was completed and results are detailed in \Cref{tab:abs_single}. The results highlight that the baseline is not effective as the feature fusion methods AUX, ADD, or MULTI. It is also evident from \Cref{tab:abs_single} that Affinity Propagation is the predominant clustering method associated with the majority of the best outcomes.

The results from \Cref{tab:abs_single} also indicate that the most common model is the FCN (ResNet50) with the MULTI feature fusion method. All scores are within 100$\mu$L and exhibit good mean hit rates. The highest mean hit rate, 66.6\%, is achieved by the FCN (EfficientNetB0) with AUX feature fusion using center points. However, this model does not record the lowest absolute difference. The smallest absolute difference, 20.9$\mu$L, is observed in the FCN (ResNet50) using MULTI feature fusion and Affinity propagation. 

Using the model with the lowest absolute difference, further analysis is carried out. \Cref{tray_coverage} includes calculations of predicted coverage, predicted deposition estimation, and comparisons to the ground truth for that particular image. These results are visually represented in \Cref{coverage_depo_stuff}.

The FCN (ResNet50) with MULTI feature fusion provides very accurate coverage rates for sprayed lettuce, within 0.02 \textit{$CM^{2}$}. The coverage for chickweed, meadowgrass, and sprayed chickweed remains quite precise, within 0.35 \textit{$CM^{2}$}, 1.26 \textit{$CM^{2}$}, and 13.27 \textit{$CM^{2}$} respectively. Visually, the coverage results in \Cref{fig3a} appear consistent across all classes. However, it is worth noting that the sprayed Meadowgrass class poses prediction challenges. Moreover, the deposition values for the lettuce and chickweed class are correct, despite not precisely identifying the key points, achieving a mean hit rate of 49.8\%, and 90.8\%, respectively. Finally, as the model does not predict sprayed meadowgrass.

\begin{table}[h]
\centering
\caption{Tray 27 FCN ResNet50 Coverage and Deposition Predictions}
\label{tray_coverage}
\begin{tabular}{lrrrrr}
Class & \begin{tabular}[r]{@{}r@{}}Actual\\  Coverage\\ ($CM^{2}$)\end{tabular} & \begin{tabular}[r]{@{}r@{}}Predicted \\ Coverage\\ ($CM^{2}$)\end{tabular} & \begin{tabular}[r]{@{}r@{}}Actual \\ Weight\\ ($\mu$L)\end{tabular} & \begin{tabular}[r]{@{}r@{}}Predicted\\ Weight\\ ($\mu$L)\end{tabular}  & \begin{tabular}[r]{@{}r@{}}Mean \\Hit \\Rate\end{tabular} \\ \hline
Lettuce & 0 & 2.89 & N/A & N/A & N/A\\ 
Chickweed & 1.14 & 0.79 & N/A & N/A & N/A\\ 
Meadowgrass & 3.65 & 4.91 & N/A & N/A & N/A\\ 
Sprayed Lettuce & 42.01 & 41.99 & 167.2 & 167.2 & 49.8\%\\ 
Sprayed Chickweed & 52.86 & 39.59 & 188.1 & 188.1 & 90.9\%\\ 
Sprayed Meadowgrass & 11.90 & 0 & 20.9 & 0 & 0\%\\ 
\end{tabular}
\end{table}

\section{Conclusion}
\label{sec:future}
This research marks an initial exploration into evaluating post-spraying effectiveness in precision agriculture without traditional tracers or WSPs for deposition quantification. We have demonstrated the capability to accurately distinguish and classify sprayed weeds and lettuces, and estimated spray weights on these targets.

Among the models evaluated, the DeeplabV3 (ResNet50) without feature fusion was the most accurate for segmenting lettuces and weeds across all metrics. While there is not a definitive model for interpretability, AblationCAM notably outperformed ScoreCAM. For precise deposition estimation, the FCN (ResNet50) with ADD feature fusion was the most effective.

The introduction of inference-only feature fusion not only enhanced interpretability but also increased accuracy within deep learning architectures. Further testing is needed to validate the efficacy of this method.

Future work is to identify individual droplets with a higher granularity in images to then compare directly to tracers or WSPs for exact deposition values. 

\bibliographystyle{IEEEtran}
\bibliography{Bibliography.bib}

\end{document}